\documentclass[letterpaper, 10 pt, conference]{ieeeconf}  

\IEEEoverridecommandlockouts                              

\makeatletter
\let\NAT@parse\undefined
\makeatother
\usepackage[numbers,sort&compress]{natbib}

\usepackage[pdftex]{graphicx}
\usepackage{amsmath}
\usepackage{amssymb}  
\usepackage{subcaption}
\usepackage{multirow}
\usepackage{array,booktabs}
\usepackage{diagbox}
\usepackage{balance}
\usepackage{xspace}
\usepackage{algorithm}
\usepackage{algpseudocode}
\usepackage{adjustbox}
\usepackage{romannum}
\usepackage{tabularx}

\usepackage[final]{hyperref}
\hypersetup{
 colorlinks=true,
 linkcolor=blue,
 filecolor=magenta,
 urlcolor=blue,
 citecolor=black
}
\usepackage{cleveref}
\crefname{figure}{Fig.}{Figs.}
\Crefname{figure}{Fig.}{Figs.}
\usepackage[font=small]{caption}
\usepackage{rpm_packages/rpm_SIunits}
\usepackage{rpm_packages/rpm_acronyms}
\usepackage{rpm_packages/rpm_math}
\usepackage{rpm_packages/rpm_misc}

\usepackage{my_packages/my_math}
\usepackage{my_packages/my_misc}
\usepackage{my_packages/my_acronyms}

\usepackage{bbm}

\usepackage{textcomp}

\usepackage{soul,color}
\usepackage{lipsum}
\usepackage{dsfont}

\usepackage{xcolor}
\usepackage{colortbl}

\DeclareMathOperator*{\argmin}{argmin}

\usepackage{amssymb}
\usepackage{pifont}
\usepackage{tikz} 
\usepackage{subcaption}
\usepackage{caption}
\title{\LARGE \bf 
Calousel: Extrinsic Calibration of Non-overlapping\\Multi-camera Systems from Pure Rotation

}     

\author{Gwanhyeong Song${}^{1}$, Chaehyeon Song${}^{1}$, and Ayoung Kim${}^{1*}$%
\thanks{$^\dagger$This work is supported by the Korea Agency for Infrastructure Technology Advancement (KAIA) grant funded by the Ministry of Land, Infrastructure and Transport (Grant RS-2023-00250727) through the Korea Floating Infrastructure Research Center at Seoul National University.}
\thanks{$^{1}$G. H. Song, C. H. Song, and A. Kim are with the Dept. of Mechanical Engineering, SNU, Seoul, S. Korea {\tt\small [skh8464, chaehyeon, ayoungk]@snu.ac.kr}}%
}

\begin{document}

\maketitle
\thispagestyle{empty}
\pagestyle{empty}

\begin{abstract}
Extrinsic calibration of multi-camera systems with non-overlapping FOVs has been a challenging problem in the robotics literature. 
Conventional target-based methods impose substantial target setup overhead, either deploying large calibration targets or requiring pre-measured multi-target poses.
Motion-based approaches instead suffer from drift error, scale ambiguity, and motion degeneracy.
Securing both accuracy and usability, we propose a novel calibration method that leverages pure rotational motion, requiring \textit{only} a single static calibration board.
The key idea is to make all cameras sequentially observe the same target under a shared geometric reference, even without overlapping views. 
To integrate these time-separated observations, we formulate the problem using a latent turntable frame and a 3D error on SE(3) within a global optimization framework.
We validate the proposed method on both a controlled camera rig and a full-scale vehicle platform with heterogeneous cameras, and analyze robustness under non-ideal turntable motion.
Extensive experiments show that our approach maintains competitive accuracy without specialized precision hardware, proving its strong suitability for realistic on-site deployments.
Our code is publicly available \href{https://github.com/Isornorphism/calousel}{here}.
\end{abstract}
\section{Introduction}
\label{sec:intro}

Multi-camera systems have become key sensing modalities for modern robotic platforms, providing wide field coverage and reliable multi-view perception~\cite{singh2023surround, wu2025omniocc}.
The performance of such platforms depends on accurate extrinsic calibration between cameras~\cite{leizea2023extrinsicpropagation}.
In practical deployments, cameras are often oriented in different directions to maximize coverage with a limited number of cameras, resulting in little to no \ac{FOV} overlap.
This makes extrinsic calibration challenging, since conventional feature-matching approaches rely on direct correspondences observed in shared views~\cite{xia2018global}.

To address non-overlapping \ac{FOV} scenarios, existing approaches can be categorized into target-based and motion-based paradigms.
Target-based methods use either a large calibration target~\cite{li2013patterns, liu2011largetarget, strauss2014calibrating} that is partially visible in each camera's view or multiple targets~\cite{liu2011multitargets, yin2018multitargets, zhu2022multitargets} placed within individual camera views.
However, the large high-precision target is often costly and cumbersome, while multi-target systems demand the precise 6D configuration data of every target for calibration.
Such setup complexity reduces practical usability, especially for repeated calibration and on-site deployment.
On the other hand, motion-based methods estimate extrinsic parameters from camera trajectories obtained via visual odometry~\cite{xu2022slam, carrera2011slam, dai2024sfmpatterns}.
Yet, these approaches are prone to drift accumulation, sensitive to motion degeneracy~\cite{guan2025affine, clipp2008robust}, and suffer from scale ambiguity without metric constraints.

\begin{figure}[!t]
  \centering
  \includegraphics[width=1\linewidth]{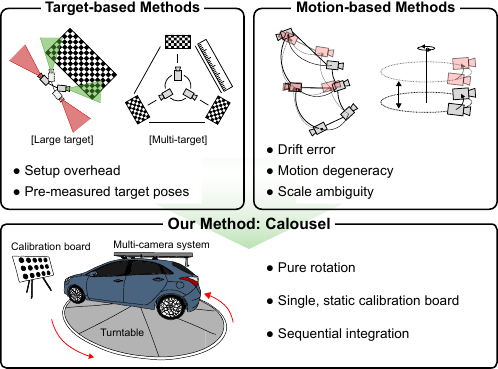}
  \caption{
  \textbf{Conceptual comparison of calibration methodologies for non-overlapping camera systems.}
  (Left) Target-based methods require large or multi-target setups, incurring setup overhead and often requiring pre-measured target poses
  (Right) Motion-based methods estimate extrinsics from trajectories but suffer from drift, motion degeneracy, and scale ambiguity.
  (Bottom) Our proposed method, \textit{Calousel}, leverages a turntable to induce space-efficient, single-axis pure rotation.
  This enables a multi-camera system (\eg, vehicle) to sequentially observe a single, static calibration board.
  By globally optimizing these views, \textit{Calousel} provides a practical balance between usability and accuracy.}
  \label{fig:overview}
  \vspace{-7mm}
\end{figure}

To balance these trade-offs, hybrid methods combine target observations with constrained motions~\cite{yang2019camera, shi2017error, tabb2017solving}. 
Although these approaches mitigate some limitations above, they do not explicitly address non-overlapping \ac{FOV}s and may not extend to extreme front–rear configurations ($\sim 180\degree$).
Moreover, they often require precisely controlled motion during data acquisition, which increases setup complexity and limits scalability to large multi-camera systems.

We present a calibration method explicitly designed for non-overlapping \ac{FOV} multi-camera systems.
Our key idea is to leverage in-place pure rotation generated by a single-axis turntable, enabling space-efficient calibration even for large-scale setups.
This motion is well suited to non-overlapping \ac{FOV} calibration because it allows all cameras to sequentially observe the same static calibration target, thereby establishing a shared metric reference.
Unlike prior hybrid pipelines that may require specialized precision motion hardware, our method uses a standard single-axis turntable, which is common in industrial workflows~\cite{jing2017sampling} and easy to replicate at a laboratory scale.
In our setup, a rigidly mounted multi-camera system rotates in front of a single static calibration board (\figref{fig:overview}).
We then integrate these time-separated observations using a latent turntable frame and a global optimization with a 3D error on $\SE3$.

We validate the proposed method on a camera rig across various configurations and experimental conditions.
We further assess scalability and real-world applicability on a full-scale vehicle platform equipped with heterogeneous cameras, by integrating rolling-shutter compensation module.
We also analyze robustness to non-ideal turntable motions (\eg, wobble and vibration), which is commonly observed in cost-effective hardware used in practical deployments.

The main contributions are summarized as follows:

\begin{itemize}
    \item We propose a space-efficient extrinsic calibration framework for non-overlapping multi-camera systems using a single-axis turntable, designed to balance practical deployability and calibration accuracy without requiring precision equipment or complex target setups.
    \item We develop a turntable-aware formulation with a latent turntable frame and a 3D error on $\SE3$ to integrate sequential target observations under pure rotation, while supporting rolling-shutter compensation.
    \item We validate the proposed method on platforms ranging from a controlled camera rig to a full-scale vehicle, and demonstrate stable calibration performance under realistic sensing and actuation artifacts.
\end{itemize}

\section{related work}
\label{sec:relatedwork}

The extrinsic parameter calibration of multi-camera systems is a critical prerequisite for accurate downstream tasks.
While overlapping \ac{FOV}s allow for direct feature matching, non-overlapping configurations leave no shared visual overlap for establishing correspondences.

\noindent\textbf{Creating Virtual Overlap}: 
One line of work attempts to virtually convert non-overlapping systems into overlapping ones.
Early approaches attempted to create virtual overlaps using optical components like mirrors~\cite{kumar2008mirror}, but they are highly susceptible to the precision of the optical equipment.
Other works~\cite{robinson2017intercamera} introduce intermediate cameras to bridge two disjoint views, yet repeated composition of transformation matrices may accumulate numerical errors, and frequently fails in extreme angular-separation configurations.

\noindent\textbf{Target-based Methods}: 
Without altering the non-overlapping conditions, \textit{target–based methods} exploit known calibration patterns to precisely estimate camera poses from image observations.
Some approaches utilize a single massive target that can be partially observed by multiple cameras~\cite{li2013patterns, liu2011largetarget, strauss2014calibrating}.
Because these observations are anchored to a unified global geometry, the extrinsic parameters can be estimated with high accuracy.
However, the applicability of a single target diminishes when the camera system features a wide baseline or large angular separation.
To overcome this, room-scale calibration environments have been proposed, enclosing the entire platform to ensure target visibility across a wide range of viewing directions~\cite{geiger2012automatic, wiesmann2024joint}.
While effective, these dedicated facilities require substantial infrastructure and careful setup.

Meanwhile, other works~\cite{liu2011multitargets, yin2018multitargets, zhu2022multitargets} locate distinct targets within each camera's \ac{FOV}.
Yet, this strategy demands the precise 6D pre-measurement of all targets, which becomes increasingly cumbersome as the sensor count grows.

\noindent\textbf{Motion-based Methods}: 
Alternatively, \textit{motion-based methods} eliminate external apparatus by estimating extrinsic parameters from visual \ac{SLAM}~\cite{xu2022slam, carrera2011slam} or \ac{SFM}~\cite{dai2024sfmpatterns}, and formulating the task as a hand-eye calibration problem.
Despite the appeal of such flexible setups, their accuracy is fundamentally bounded by the quality of visual odometry and can degrade under drift.
Moreover, hand-eye calibration is highly sensitive to motion degeneracy (\eg, pure rotation or translation~\cite{guan2025affine, clipp2008robust}).
Monocular setups also suffer from inherent scale ambiguity, necessitating supplementary metric cues such as IMU measurements~\cite{rehder2016otherdevice}, range sensors or references to known objects, which increase hardware dependency and environmental sensitivity.

\noindent\textbf{Hybrid Strategies}: Consequently, \textit{hybrid strategies} combining target observations with controlled motion have emerged as a solution.
For example, prior works combined a calibration target with controlled motions generated by pure translation stages~\cite{yang2019camera}, two-axis turntables~\cite{shi2017error}, and robot arms~\cite{tabb2017solving}.
Despite their promise, many hybrid methods are not explicitly designed for strictly non-overlapping \ac{FOV} systems.
Moreover, precision hardware and alignment procedures can increase setup complexity and cost, thereby limiting scalability to large-scale systems and extreme angular-separation configurations.

To bridge these gaps, we leverage a single-axis turntable to rotate the multi-camera system in place.
This allows cameras with non-overlapping \ac{FOV}s, even in front-rear configurations, to sequentially observe a single static calibration board, providing a space-efficient alternative to complex multi-axis actuators.
Crucially, this setup inherently resolves two fundamental limitations: the board provides metric constraints that remove scale ambiguity, and camera-to-board absolute poses avoid degeneracy under pure rotation.
By globally optimizing these sequential observations with a 3D error on $\SE3$, we achieve robust, competitive accuracy.
\section{Method}
\label{sec:method}
This section presents a pipeline for accurate extrinsic calibration of a multi-camera system $\{C_i\}_{i=0}^{N-1}$, specifically targeting configurations with non-overlapping \ac{FOV}. The method is structured into three stages. First, we define the problem setup and coordinate frames, and describe keyframe selection (including rolling-shutter compensation when applicable).
Second, we derive robust kinematics-based initialization for the turntable and camera mounting parameters.
Lastly, we refine all parameters jointly via a global optimization that minimizes a 3D error function.

\subsection{System Setup and Coordinate Frames Definition}
\label{sec:preliminaries}

\begin{figure}[!b]
  \centering
  \includegraphics[width=1\linewidth]{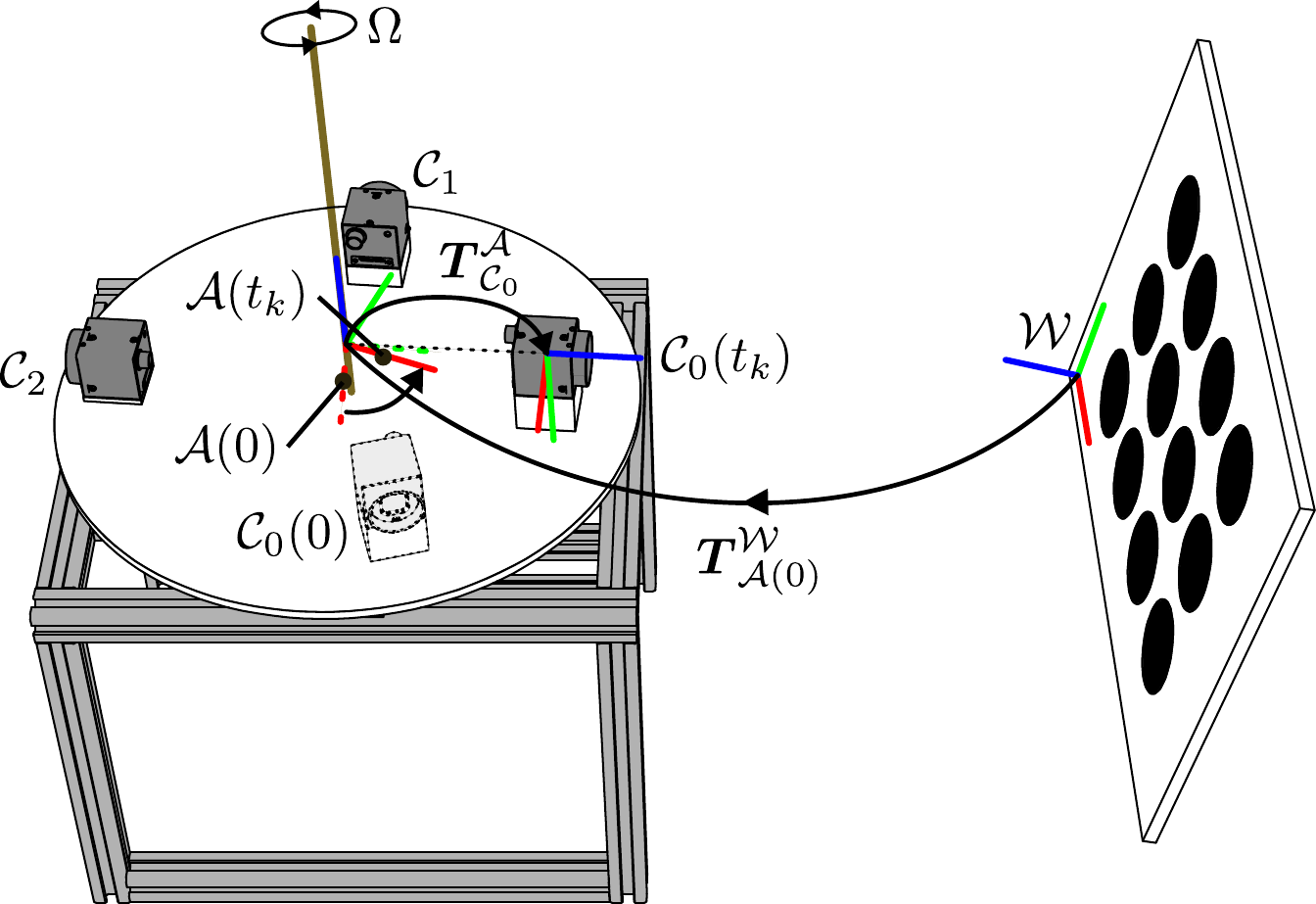}
  \caption{\textbf{Coordinate frame setting of the proposed calibration system.} The world frame $\coord{W}$ is fixed to the calibration board. The turntable frame $\coord{A}(t)$ rotates with the multi-camera system at a constant angular velocity $\Omega$. $\coord{C}_i$ denotes the frame of the $i$-th camera, with $\coord{C}_0$ being the reference camera. The coordinate axes are coded as red for the $x$-axis, green for the $y$-axis, and blue for the $z$-axis.}
  
  \label{fig:frame_setting}
  \vspace{-4mm}
\end{figure}

Our methodology is implemented using a calibration board fixed in the environment and a rotary turntable that applies pure rotational motion to the multi-cameras system. As a prerequisite, we assume the intrinsic parameters of each camera have been pre-calibrated using the method from \cite{song2024unbiased}.

For clarity, we represent an $\SE3$ transformation $\mat{T}$ by its rotation matrix and translation vector as $\mat{T} = (\mat{R},\bvec{t})$. Throughout the paper, scalar $t$ denotes time, and boldface denotes vectors.

Our algorithm uses the following physical assumptions:

\begin{itemize}
    \item The multi-camera system, rigidly mounted on the turntable, is an ideal rigid body that undergoes no structural deformation during motion.
    \item The turntable executes pure rotation with a constant angular velocity $\Omega$.
    \item During data acquisition, the turntable completes at least one full rotation, and every camera is guaranteed to observe the calibration board at a certain interval.
\end{itemize}
The second assumption defines an idealized motion model. Since real turntables can exhibit non-ideal effects such as axis wobble and vibration, we later analyze such deviations and their impact on the estimated extrinsics in \secref{sec:axis_deviation}.

As shown in \figref{fig:frame_setting}, we define three coordinate frames.
The world frame $\coord{W}$ is attached to the calibration board, with its $xy$-plane coincides with the board plane.
Each camera frame is denoted $\coord{C}_i(t)$, whose $z$-axis aligns with the principal axis (\ie, viewing direction).
We further introduce a latent turntable frame $\coord{A}(t)$, which is fixed to the turntable and rotates about its $z$-axis over time.
We term $\coord{A}(t)$ \textit{latent} because it is not directly observed from image measurements; instead, it is introduced to separate the turntable-driven time-varying rotation from the time-invariant camera mounting geometry in the calibration.
The efficacy of this representation is further examined in \secref{sec:ablation}.

The turntable frame $\coord{A}(t)$ admits gauge ambiguities: the in-plane orientation of its $x$ and $y$ axes is ambiguous over time, and its origin is also ambiguous along the rotation axis. To fix these \ac{DOF}, we impose two constraints. First, at the time origin $t=0$, we choose $\coord{A}(0)$ such that its $x$-axis is as closely aligned with the $x$-axis of $\coord{W}$ as possible. Second, we set the origin of $\coord{A}(t)$ so that the reference camera $C_0$ lies on the $z=0$ plane in $\coord{A}(t)$. The definitions of the time origin and the reference camera are provided in \secref{sec:keyframe_selection}.

With this construction, two transformations are nominally time-invariant: the turntable pose at the time origin, $\pose{\mat{T}}{\coord{W}}{\coord{A}(0)}$, and the mounting pose of each camera relative to the turntable, $\pose{\mat{T}}{\coord{A}}{\coord{C}_i}$. 
Hence, we omit the time index in $\pose{\mat{T}}{\coord{A}}{\coord{C}_i}$ (\ie, $\pose{\mat{T}}{\coord{A}(t)}{\coord{C}_i(t)} \equiv \pose{\mat{T}}{\coord{A}}{\coord{C}_i}$).

Based on these definitions, the kinematic model for the predicted pose of camera $C_i$ at timestamp $t_k$ is modeled as 
\begin{equation}
    \pose{\mat{\hat{T}}}{\coord{W}}{\coord{C}_i(t_k)} = \pose{\mat{T}}{\coord{W}}{\coord{A}(0)} \mathrm{rot}(\hat{z}, \; \Omega t_k) \pose{\mat{T}}{\coord{A}}{\coord{C}_i},
    \label{eq:kinematics}
\end{equation}
where $\mathrm{rot}(\hat{z}, \; \theta)$ denotes a pure rotational transformation by angle $\theta$ about the $z$-axis in $\SE3$. The unknown parameters consist of the angular velocity $\Omega$ and the turntable pose at the time origin, $\pose{\mat{T}}{\coord{W}}{\coord{A}(0)}$, and the extrinsic poses, $\pose{\mat{T}}{\coord{A}}{\coord{C}_i}$ for $i=0,\dots,(N-1)$. These parameters collectively have $(6N-1)$ \ac{DOF} when imposing the constraint $\bvec{t}_z=0$ for the reference camera, $C_0$. Consequently, the complete parameter set optimized in this study is $\{\Omega, \pose{\mat{T}}{\coord{W}}{\coord{A}(0)}, \pose{\mat{T}}{\coord{A}}{\coord{C}_i} (i=0,\dots,N-1) \}$, comprising $(6N+6)$ parameters.

\subsection{Keyframe Selection and Rolling-Shutter Compensation}
\label{sec:keyframe_selection}

\noindent{\textbf{Keyframe Selection}}: To reduce the computational load, our algorithm selects keyframes instead of using all images. For each camera $C_i$, we process a sliding window of consecutive images in which the calibration board is visible within the \ac{FOV}. Within each window, we perform target-based camera pose estimation using \cite{song2024unbiased}, and we append the frame with the lowest reprojection error to a dedicated keyframe queue for that camera. Each selected keyframe stores the estimated camera-to-board pose and its associated covariance, $\Sigma_{\xi}$, defined in $\se3$. To avoid near-duplicate viewpoints and improve numerical conditioning, we adapt the sampling such that the rotation between any two adjacent selected keyframes exceeds a threshold $\theta_{th}$. This yields a per-camera keyframe sequence ${(t_k, \pose{\mat{T}}{\coord{W}}{\coord{C}_i(t_k)}, \Sigma_{\xi(t_k)}})_k$, where the index $k$ is local to each camera.

Now, we define the \textit{time origin} $t=0$ to the timestamp of the globally best keyframe (the keyframe achieving the minimum reprojection error across all cameras and windows), and we designate the camera that captured it as the \textit{reference camera}, which we re-index as $C_0$ without loss of generality.

\noindent{\textbf{Rolling-Shutter Compensation}}: Rolling-shutter cameras can introduce noticeable bias in target-based pose estimation under rotational motion~\cite{oth2013rolling}, because image rows are exposed at different times within a single frame. To mitigate this effect, we incorporate a lightweight rolling-shutter compensation module as a first-order approximation.

Within each sliding window, we estimate the average pixel velocity of the detected circle centroids, $(\dot{u}, \dot{v})$, by comparing the centroid locations between the first and last successful detections in the window and dividing by the time difference. Let $H$ be the image height in pixels and $t_{\mathrm{readout}}$ the total sensor readout time per frame. Assuing a uniform row-wise readout, the exposure time offset for a measurement at image row $r$ is approximated as $t_{\mathrm{line}} = t_{\mathrm{readout}}(r/H)$. We then compensate the predicted projection $(u_e, v_e)$ by $(u_e, v_e) \leftarrow (u_e+t_{\mathrm{line}}\dot{u}, v_e + t_{\mathrm{line}}\dot{v})$ before computing residuals.

In practice, we re-run the target-based pose estimation for rolling-shutter keyframes by augmenting the optimization with $t_{\mathrm{readout}}$ and integrating the above correction into the residual computation. A qualitative evaluation of the resulting improvement is provided in \secref{sec:vehicle_eval}.

\subsection{Turntable-Kinematics-Based Initialization}
\label{sec:initialization}
\noindent \textbf{Initialization of Rotational Components}:
Under the constant-velocity rotation assumption, relative rotations observed from all cameras should share a common rotation axis. We initialize the turntable motion per revolution (cycle) by using relative rotations with respect to the first keyframe of that cycle.

Specifically, let $t_0$ denote the timestamp of the first keyframe in a cycle and $t_k$ that of a subsequent keyframe. The relative rotation between the two poses is mapped to the Lie algebra to extract the rotation angle and unit axis as 

\begin{equation}
    \bvec{\hat{u}}_k\Delta\theta_k = \Log \left( \pose{\mat{R}}{\coord{W}}{\coord{C}_i(t_k)} (\pose{\mat{R}}{\coord{W}}{\coord{C}_i(t_{0})})^{-1}\right),
\end{equation}
Here, $\Delta\theta_k$ is the rotation angle and $\bvec{\hat{u}}_k$ is the unit rotation axis. An angular velocity estimate is obtained as $\Omega = \Delta\theta_k / (t_k-t_0)$. These estimates are computed for all keyframes within the cycle, the angular velocity estimates are averaged to obtain the initial angular velocity, $\Omega$, and the axis vectors are averaged and normalized to obtain the initial rotation axis, $\bvec{\hat{u}}$.


\noindent \textbf{Initialization of the Origin of Turntable Frame}: A reliable initialization of the turntable-frame origin is crucial, as it affects not only the initial estimate of the turntable pose  $\pose{\mat{T}}{\coord{W}}{\coord{A}(0)}$, but also the initial estimates of the camera mounting poses $\pose{\mat{T}}{\coord{A}}{\coord{C}_i}$. In practice, each camera acquires camera-to-board poses only over the angular range where the calibration board lies within its \ac{FOV}. Hence, camera trajectories are only partially observed over a full revolution, and the observable range can be particularly limited along the world $z$-direction. Under these constraints, we propose a method to robustly estimate the turntable origin, $\pose{\bvec{t}}{\coord{W}}{\coord{A}(0)}$, by solving a linear least-squares problem that aggregates geometric constraints from the available trajectories. 

The origin estimation combines two complementary conditions with different scopes of applicability. First, we impose an \textit{equidistance condition} for \textit{every camera trajectory}. Since each camera is rigidly mounted on the rotating platform, the distance between the turntable origin and the camera center must remain constant over time. For two time stamps $t_{0}, t_{k}$ of the same camera $C_i$, this condition can be linearized as shown in \eqref{eq:equidistance}. For $K_i$ keyframes of camera $C_i$, this yields $(K_i- 1)$ linear constraints.

{\small
\begin{align}
    & \norm{\pose{\bvec{t}}{\coord{W}}{\coord{C}_i(t_{0})} - \pose{\bvec{t}}{\coord{W}}{\coord{A}(0)}} = \norm{(\pose{\bvec{t}}{\coord{W}}{\coord{C}_i(t_{k})} - \pose{\bvec{t}}{\coord{W}}{\coord{A}(0)}}, \\
    & 2\left(\pose{\bvec{t}}{\coord{W}}{\coord{C}_i(t_{k})} - \pose{\bvec{t}}{\coord{W}}{\coord{C}_i(t_{0})} \right)^\top \pose{\bvec{t}}{\coord{W}}{\coord{A}(0)} = \norm{\pose{\bvec{t}}{\coord{W}}{\coord{C}_i(t_{k})}}^2 - \norm{\pose{\bvec{t}}{\coord{W}}{\coord{C}_i(t_{0})}}^2.
    \label{eq:equidistance}
\end{align}
}

Second, we impose a \textit{coplanarity condition} only for \textit{reference camera} $C_0$. Recall that the turntable frame $\coord{A}$ is defined such that the reference camera lies on its $z=0$ plane. Equivalently, the relative displacement from the reference camera center to the turntable origin, $\pose{\bvec{t}}{\coord{W}}{\coord{A}(0)} - \pose{\bvec{t}}{\coord{W}}{\coord{C}_0(t_k)}$, must be orthogonal to the turntable axis $\bvec{\hat{u}}$, yielding the coplanarity (axis-orthogonality) constraint:
\begin{align}
    & \bvec{\hat{u}}^\top \left(\pose{\bvec{t}}{\coord{W}}{\coord{A}(0)} - \pose{\bvec{t}}{\coord{W}}{\coord{C}_0(t_k)} \right) = 0, \\
    & \bvec{\hat{u}}^\top \pose{\bvec{t}}{\coord{W}}{\coord{A}(0)} = \bvec{\hat{u}}^\top \pose{\bvec{t}}{\coord{W}}{\coord{C}_0(t_k)}.
    \label{eq:coplanarity}
\end{align}

We stack all linear equations from \eqref{eq:equidistance} (all cameras) and \eqref{eq:coplanarity} (reference camera only) into a single system $\mat{A}\bvec{x} = \bvec{b}$, where $\mat{A} \in \mathbb{R}^{(K-N+K_0) \times 3}$ with $K=\sum_i K_i$. The initial estimate $\pose{\bvec{t}}{\coord{W}}{\coord{A}(0)}$ is obtained by solving this linear least squares.

\begin{equation}
    \pose{\bvec{t}}{\coord{W}}{\coord{A}(0)} = \argmin_{\pose{\bvec{t}}{\coord{W}}{\coord{A}(0)}} \norm {\bvec{b}-\mat{A} \pose{\bvec{t}}{\coord{W}}{\coord{A}(0)}}^2.
\end{equation}


\noindent \textbf{Initialization of the Camera Mounting Pose}: Given the initial turntable parameters $\{\Omega, \pose{\mat{T}}{W}{A(0)}\}$ and the observed camera-to-board poses $\pose{\mat{T}}{\coord{W}}{\coord{C}_i(t_k)}$, we compute candidates for each camera mounting pose $\pose{\mat{T}}{\coord{A}}{\coord{C}_i}$ by rearanging the kinematic model in \eqref{eq:kinematics}. Since $\pose{\mat{T}}{\coord{A}}{\coord{C}_i}$ is modeled as a time-invariant variable shared across all keyframes of camera $C_i$, we obtain a robust initial estimate by averaging these candidate in $\se3$ Lie algebra and mapping back to $\SE3$ via $\Exp(\cdot)$.


\subsection{Global Optimization using 3D Error Function}
\label{sec:BA}

To estimate the accurate extrinsic parameters of the camera system, $\pose{\mat{T}}{\coord{A}}{\coord{C}_i}$, we simultaneously optimize all system parameters. In target-based calibration, the conventional cost function is the reprojection error defined in the 2D image space. However, our optimization process directly minimizes \eqref{eq:RPE}, the pose error between the observed camera pose, $\pose{\mat{T}}{\coord{W}}{\coord{C}_i}$, and the predicted camera pose $\pose{\mat{\hat{T}}}{\coord{W}}{\coord{C}_i}$, calculated from the kinematic model in \eqref{eq:kinematics}. 

\begin{align}
    & \{\Omega,\pose{\mat{T}}{\coord{W}}{\coord{A}(0)},\pose{\mat{T}}{\coord{A}}{\coord{C}_i} (i=0,\dots,N-1)\} \nonumber\\
    & = \argmin_{\Omega,\pose{\mat{T}}{\coord{W}}{\coord{A}(0)},\pose{\mat{T}}{\coord{A}}{\coord{C}_i}} \sum_{i, k} \norm{ \Log \left( (\pose{\mat{\hat{T}}}{\coord{W}}{\coord{C}_i(t_k)})^{-1} \pose{\mat{T}}{\coord{W}}{\coord{C}_i(t_k)}\right)}_{\Sigma}^2.
    \label{eq:RPE}
\end{align}
Here, $\norm{\cdot}_\Sigma$ denotes the Mahalanobis norm with respect to measurement covariance $\Sigma_{\bvec{\xi}}$ defined in $\se3$.
This strategy mitigates the tendency of reprojection-based optimization to excessively reduce the error of individual cameras. A detailed accuracy comparison with reprojection error–based optimization can be found in \secref{sec:ablation}.

We empirically found that the strong kinematic constraints of the turntable system induce high correlations between the components of the $\se3$ vector, leading to an optimization instability. To mitigate this, we simplify the weighting by using only the diagonal elements of the covariance matrix.
Minimizing the loss function in \eqref{eq:RPE} with this weighting scheme yields the final extrinsic parameters $\pose{\mat{T}}{\coord{A}}{\coord{C}_i}\; (i=0,\dots,N-1)$, which are defined with respect to the turntable body-fixed frame $\coord{A}(t)$.

\section{experiment}
\label{sec:experiment}






This section evaluates the proposed extrinsic calibration method on two platforms: a robotic camera rig and a full-scale vehicle.
Calibration accuracy is first quantified on the robotic camera rig under diverse non-overlapping \ac{FOV} configurations, which serves as a compact sensor setup targeting small robotic platforms.
The same evaluation is then extended to the full-scale vehicle to assess real-scale performance.
We additionally validate the rationale behind our key design choices, including the latent frame and the 3D error function in \eqref{eq:RPE} and examine the effect of rolling-shutter compensation.
Motion imperfections, including small deviations of the rotation axis and fluctuations of the turntable-frame origin, are also quantified and related to the calibration accuracy.
Finally, as an additional reference, we compare our estimates with a conventional target-based calibration method on an overlapping \ac{FOV} setting.


\subsection{Experimental Platforms and Evaluation Metric}
\label{sec:exp_explain}

\begin{figure}[t]
  \centering
  \includegraphics[width=1\linewidth]{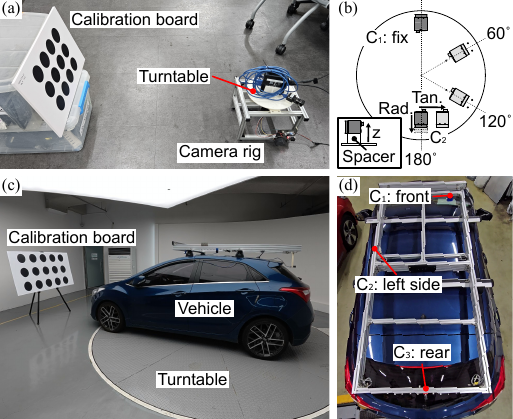}
  \vspace{-5mm}
  \caption{\textbf{Experiment setup.} (a) Benchtop camera rig including the turntable, the 3D-printed reconfigurable rig, and the calibration board.
  (b) Schematic of the reconfigurable rig illustrating angular and translational configuration sweeps used for evaluation.
  (c) Full-scale vehicle platform on an automotive turntable alongside the calibration board.
  (d) Top-down view of the vehicle roof rack, detailing non-overlapping mounting positions of the three cameras.}
  \label{fig:total_setup}
  \vspace{-3mm}
\end{figure}

\noindent \textbf{Experimental Platforms}: We collected data on two platforms, a benchtop camera rig and a full-scale vehicle, to evaluate the proposed method under both systematic configuration changes and more realistic conditions.
The camera rig uses a 3D printed reconfigurable rig that enables diverse camera configurations, while the full-scale vehicle platform is constructed with a larger and heterogeneous camera setup to assess applicability.
A summary of the hardware setups and specifications for both platforms is provided in \figref{fig:total_setup} and \tabref{tab:setup_summary}.
Across all experiments, the turntable was rotated for at least two full cycles, and the same minimum rotation threshold for keyframe selection, $\theta_{th} = 3^\degree$, was used.

For clarity, the experiments in \secref{sec:testbed_eval} and \secref{sec:compare_overlap} were conducted on the camera rig, whereas \secref{sec:vehicle_eval} reports results obtained on the full-scale vehicle.
In \secref{sec:ablation}, the rolling-shutter compensation results are obtained on the full-scale vehicle, while the remaining results of design-choice validation are evaluated on the camera rig.
Finally, \secref{sec:axis_deviation} characterizes the non-ideal turntable motion observed on both platforms.

\noindent \textbf{Evaluation Metric}: For rotation components, we use nominal relative rotations between cameras as ground truth. 
Specifically, we set the relative rotation of each camera $C_i$ with respect to $C_1$ (for $i \geq 2$), and quantify the error using Euler angle components and \ac{RRE}. 
However, establishing an absolute ground truth for translation is challenging, as the internal camera centers are not directly observable even when mounted on the designed system. 
Therefore, following the available ground-truth strategy in~\cite{jeong2019evalmethod}, we apply a known relative displacement between two configuration trials and evaluate how closely the calibrated displacement recovers this value. 
The translation error is reported as a component-wise vector in (radial, tangential, \(z\)) and by its norm (\ie, \ac{RTE}). 
We report the camera rig results in \secref{sec:testbed_eval} and the full-scale vehicle results in \secref{sec:vehicle_eval} under the same protocol.

\newcolumntype{L}[1]{>{\raggedright\arraybackslash}p{#1}}
\newcolumntype{Y}{>{\raggedright\arraybackslash}X}
\newcolumntype{Z}{>{\hsize=1.05\hsize\raggedright\arraybackslash}X}

\begin{table}[!t]
\centering

\caption{Summary of experimental platforms for the camera rig and full scale vehicle evaluations. Camera marked with [RS] uses a rolling-shutter sensor; without [RS] uses a global-shutter sensor.}
\label{tab:setup_summary}
\vspace{-2mm}
\resizebox{\columnwidth}{!}{
\renewcommand{\arraystretch}{1.4}
\begin{tabularx}{\columnwidth}{L{0.21\columnwidth}YZ}
\hline

\textbf{Item} & \textbf{Robotic camera rig} & \textbf{Full-scale vehicle} \\

\hline

\# Cameras & 2 & 3 \\
Camera types & 2$\times$ FLIR Blackfly S & \celltab{t}{l}{2$\times$ FLIR Blackfly S \\ 1$\times$ RealSense D435i [RS]} \\
Resolution & 1440$\times$1080 (FLIR) & \celltab{t}{l}{1440$\times$1080 (FLIR) \\ 1080$\times$720 (D435i)} \\
FPS & 60 & 30 \\
Turntable type & Stepper-motor turntable & Automotive turntable \\
Board config.
& \celltab{t}{l}{$4\times3$ \\ $r{=}0.035\,\m$, $d{=}0.09\,\m$}
& \celltab{t}{l}{$5\times3$ \\ $r{=}0.1\,\m$, $d{=}0.3\,\m$} \\

\hline
\end{tabularx}
}
\vspace{-3mm}
\end{table}


\subsection{Robotic Camera Rig Evaluation}
\label{sec:testbed_eval}

\tabref{tab:extrinsic_parameter_rot} and \tabref{tab:extrinsic_parameter_trans} summarize the rotation and translation results on the camera rig under multiple configurations with non-overlapping \ac{FOV}s. 
The proposed method maintains stable accuracy, with \ac{RRE} below $2.0\degree$ and \ac{RTE} below $2.0\,\mm$. 
In addition, the Euler components show consistent standard deviations (typically $0.2\degree$ to $0.5\degree$), indicating repeatable rotation estimation across different orientations.
For translation, it exhibits relatively larger mean error and variance in the radial direction, reflecting the inherent geometric uncertainty of depth estimation from 2D image observations.

\begin{table}[!t]
\centering

\caption{Comparison of estimated extrinsic rotations against ground truth on the camera rig.
Results are analyzed by Euler angle components and RRE, formatted as `mean$\pm$SD' (this notation is maintained in all subsequent tables).
All units are in $\deg$.}
\label{tab:extrinsic_parameter_rot}
\vspace{-2mm}
\resizebox{\columnwidth}{!}{
\renewcommand{\arraystretch}{1.4}
\begin{tabular}{ccc|ccc|c}
\hline
\multicolumn{3}{c|}{\textbf{GT}}
& \multicolumn{3}{c|}{\textbf{Measured}}
& \multirow{2}{*}{RRE} \\

\cline{1-6}

Roll & Pitch & Yaw
& Roll & Pitch & Yaw
& \\

\hline

0 & 0 & 60
& 0.7$\pm$0.4 & 0.3$\pm$0.2 & 60.3$\pm$0.2
& 0.9$\pm$0.2 \\

0 & 0 & 120
& 0.9$\pm$0.2 & 1.2$\pm$0.2 & 120.0$\pm$0.2
& 1.5$\pm$0.3 \\

0 & 0 & 180
& -0.1$\pm$0.2 & 1.8$\pm$0.5 & 179.8$\pm$0.5
& 1.9$\pm$0.5 \\

\hline
\end{tabular}
}
\vspace{-2mm}
\end{table}
\begin{table}[!t]
\centering

\caption{Comparison of estimated extrinsic translations against available ground truth, based on known relative displacements between two configurations, on the camera rig. Results are expressed in (radial, tangential, $z$) and RTE.
All units are in $\mm$.}
\vspace{-2mm}
\label{tab:extrinsic_parameter_trans}

\resizebox{\columnwidth}{!}{
\renewcommand{\arraystretch}{1.4}
\begin{tabular}{ccc|ccc|c}
\hline
\multicolumn{3}{c|}{\textbf{Displacement GT}}
& \multicolumn{3}{c|}{\textbf{Measured Displacement}}
& \multirow{2}{*}{RTE} \\

\cline{1-6}

Rad. & Tan. & $z$
& Rad. & Tan. & $z$
& \\

\hline
-20 & 0 & 0
& -21.0$\pm$1.2 & -0.2$\pm$1.1 & -0.4$\pm$0.5
& 2.0$\pm$0.2 \\

0 & 50 & 0
& 0.3$\pm$1.3 & 50.5$\pm$0.3 & 0.2$\pm$0.6
& 1.5$\pm$0.6 \\

0 & 0 & 10
& 0.8$\pm$1.5 & -0.2$\pm$1.2 & 9.9$\pm$0.5
& 1.8$\pm$1.1 \\

\hline
\end{tabular}
}
\vspace{-5mm}
\end{table}

Although direct comparison is difficult due to differences in sensor setups and calibration environments, our results suggest that the proposed method occupies a practical middle ground in the existing literature.
For reference, existing target-based method~\cite{yin2018multitargets} typically reports rotation errors around $0.06\degree$ and translation errors around $0.08\,\mm$.
In contrast, the motion-based method~\cite{xu2022slam} shows rotation errors of approximately $2.5\degree$ and significantly larger translation errors around $14\,\mm$.
Compared to these prior reports, while our approach does not reach the sub-millimeter precision of complex target-based systems\footnote{\eg, mutli-target setups that require target poses the 6D poses of all targets to be known \emph{a priori}.}, it compares favorably to motion-based methods.
This improvement is particularly prominent in translation (\ie, $<2.0\,\mm$ vs. $14\,\mm$), suggesting that our framework successfully mitigates the scale ambiguity and drift that typically degrade motion-based calibration.

\subsection{Full-scale Vehicle Evaluation}
\label{sec:vehicle_eval}

\tabref{tab:extrinsic_calibration_rot_real} and \tabref{tab:extrinsic_calibration_trans_real} compare the estimated extrinsic parameters of the side camera $C_2$ and the rear camera $C_3$ with respect to the front camera $C_1$ against the ground truth on the full-scale platform.
In \tabref{tab:extrinsic_calibration_rot_real}, the rotation errors remain at a similar level to those observed on the camera rig (\tabref{tab:extrinsic_parameter_rot}), indicating that the proposed method preserves rotation accuracy when scaling up to a vehicle-sized platform.
In contrast, \tabref{tab:extrinsic_calibration_trans_real} shows that translation errors are noticeably larger than the camera rig results (\tabref{tab:extrinsic_parameter_trans}), which is consistent with the larger platform scale.

\begin{figure}[!t]
  \centering
  \includegraphics[width=1\linewidth]{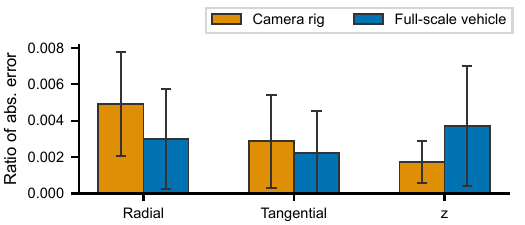}
  \vspace{-6mm}
  \caption{\textbf{Comparison of camera rig and full-scale vehicle translation errors using normalized metrics.} The plot summarizes component-wise translation error ratios across the configuration sweep for each platform, where each translation error is normalized by the corresponding inter-camera distance.
  For the camera rig, the statistics are computed from the estimated pose of $C_2$ with respect to $C_1$.
  For the full-scale vehicle, the statistics aggregate the estimated poses of both $C_2$ and $C_3$ with respect to $C_1$. Bar plots indicate \ac{MAE}, error bars denote 1-sigma.}
  \label{fig:vehicle_barplot}
  \vspace{-5mm}
\end{figure}

This behavior is expected from the lever-arm effect in error propagation~\cite{nguyen2018covariance, barfoot2014associating}.
In calibration, translation accuracy is coupled to rotational uncertainty, and even comparable rotational uncertainty can induce larger translation errors as the distance between the two frames increases.
To account for this scale dependence, \figref{fig:vehicle_barplot} reports the component-wise translation error as a ratio, obtained by dividing the absolute translation error by the inter-camera distance.
With this normalization, the camera rig and the full-scale vehicle exhibit comparable error levels across components, supporting the performance of consistence on scale.
Overall, these results on a vehicle-scale platform using an automotive turntable support the practical applicability of the proposed method to realistic industrial calibration workflows.

\begin{table}[!t]
\centering

\caption{
Comparison of estimated extrinsic rotations against ground truth on the full-scale vehicle.
$C_2$ and $C_3$ denote the extrinsic parameters relative to $C_1$
Results are summarized by Euler angle components and RRE. All units are in $\deg$.
}
\vspace{-2mm}
\label{tab:extrinsic_calibration_rot_real}
\resizebox{\columnwidth}{!}{
\renewcommand{\arraystretch}{1.4}
\begin{tabular}{c|ccc|ccc|c}

\hline

\multirow{2}{*}{\textbf{Cam}}
& \multicolumn{3}{c|}{\textbf{GT}}
& \multicolumn{3}{c|}{\textbf{Measured}}
& \multirow{2}{*}{RRE}
\\

\cline{2-7}

& Roll & Pitch & Yaw
& Roll & Pitch & Yaw
& 
\\

\hline

\multirow{2}{*}{$C_2$}
& 0 & 0 & 60
& 0.5 & 0.7 & 59.8
& 0.9
\\

& 0 & 0 & 90
& 0.3 & 0.1 & 90.6
& 1.0
\\

\hline

$C_3$
& 0 & 0 & 180
& -0.2 & -0.5 & 179.1
& 1.5
\\




\hline
\end{tabular}
}
\vspace{-2mm}
\end{table}








\begin{table}[!t]
\centering

\caption{
Comparison of estimated extrinsic translations against available ground truth on the full-scale vehicle.
$C_2$ and $C_3$ denote the extrinsic parameters relative to $C_1$.
Results are expressed in (radial, tangential, $z$) and RTE.
All units are in $\mm$.
}
\vspace{-2mm}
\label{tab:extrinsic_calibration_trans_real}
\resizebox{\columnwidth}{!}{
\renewcommand{\arraystretch}{1.4}
\begin{tabular}{c|ccc|ccc|c}

\hline

\multirow{2}{*}{\textbf{Cam}}
& \multicolumn{3}{c|}{\textbf{Displacement GT}}
& \multicolumn{3}{c|}{\textbf{Measured Displacement}}
& \multirow{2}{*}{RTE}
\\

\cline{2-7}

& Rad. & Tan. & $z$
& Rad. & Tan. & $z$
& 
\\

\hline








\multirow{2}{*}{$C_2$}
& 0 & 0 & 0
& 1.1 & 0.6 & 4.2
& 8.3
\\

& 0 & 150 & 0
& 4.0 & 151.5 & 3.6
& 5.6 
\\


\hline

\multirow{3}{*}{$C_3$}
& 0 & 0 & 0
& 1.8 & -0.8 & 2.1
& 7.7
\\

& 0 & 150 & 0
& 0.9 & 145.9 & 5.2
& 7.8
\\

& 0 & 0 & -100
& 0.6 & 0.0 & -107.8
& 11.7
\\

\hline
\end{tabular}
}
\vspace{-3.5mm}
\end{table}















\subsection{Validation of Key Method Components}
\label{sec:ablation}

\begin{figure}[t]
  \centering
  \begin{subfigure}[t]{0.49\linewidth}
    \centering
    \begin{tikzpicture}
      \node[inner sep=0] (img) {\includegraphics[width=\linewidth]{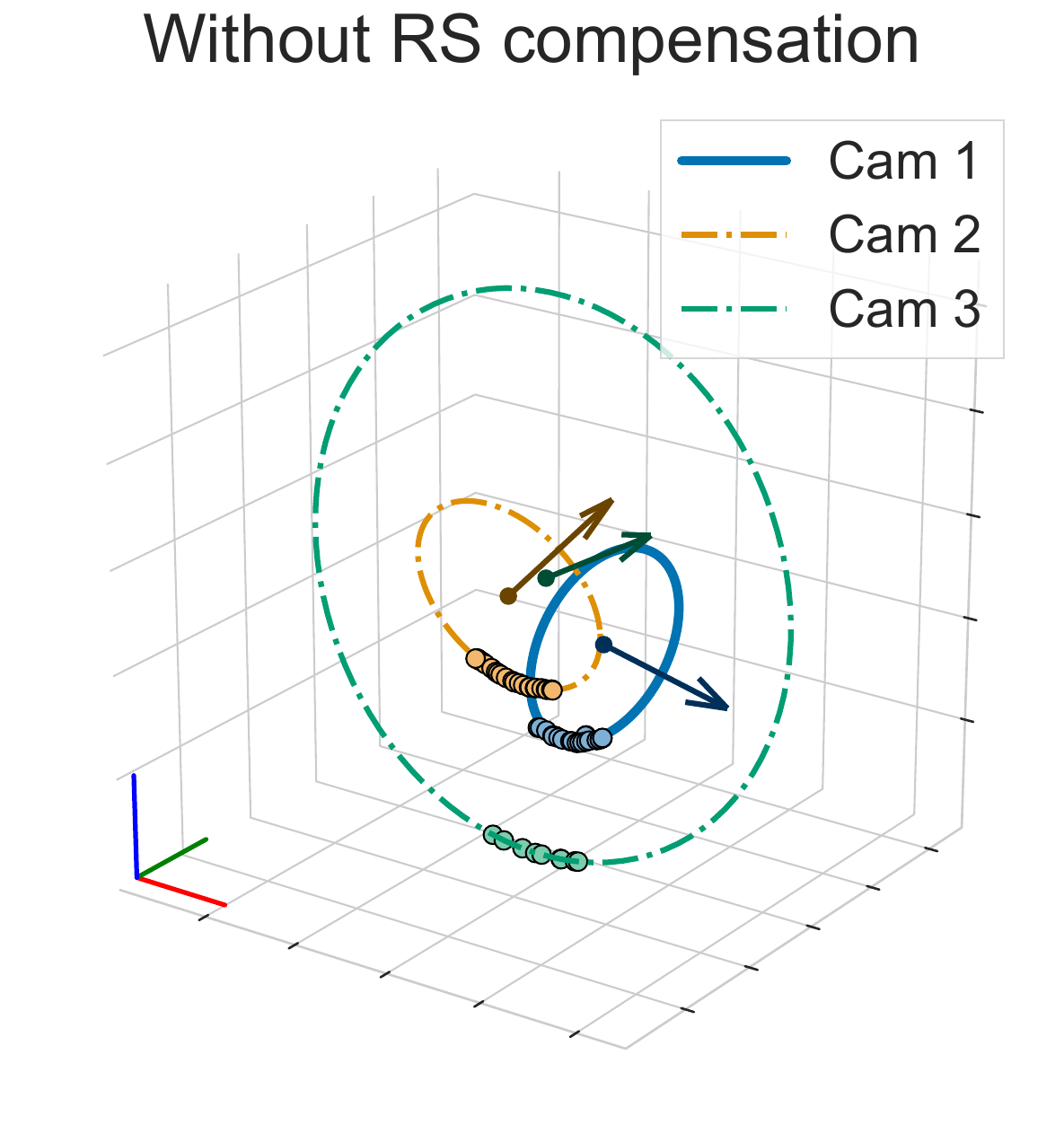}};
      \node[anchor=north west, fill=white, draw=none, inner sep=1pt] at (img.north west) {\footnotesize(a)};
    \end{tikzpicture}
  \end{subfigure}
  \hfill
  \begin{subfigure}[t]{0.49\linewidth}
    \centering
    \begin{tikzpicture}
      \node[inner sep=0] (img) {\includegraphics[width=\linewidth]{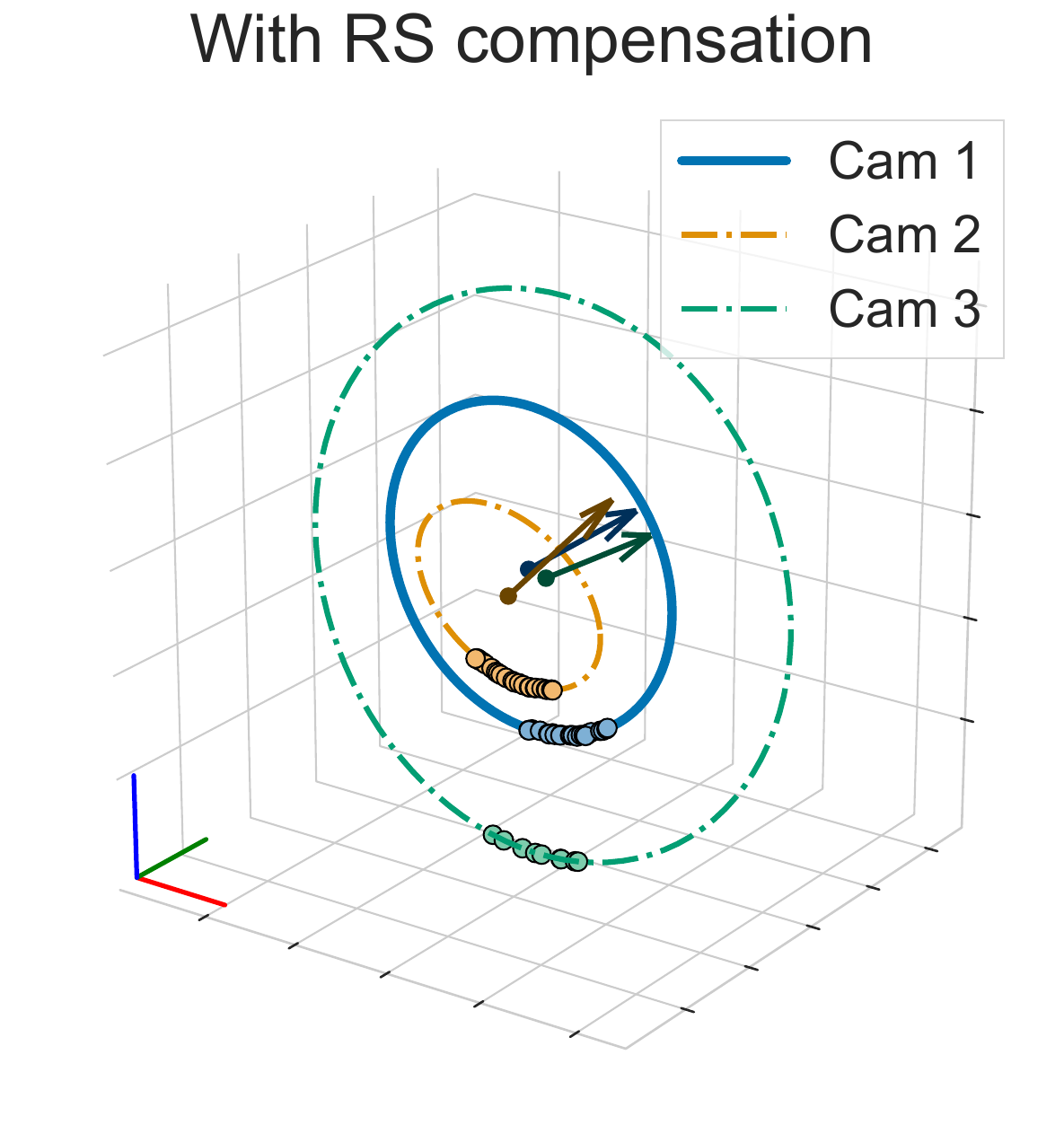}};
      \node[anchor=north west, fill=none, draw=none, inner sep=1pt] at (img.north west) {\footnotesize(b)};
    \end{tikzpicture}
  \end{subfigure}
  \vspace{-6mm}

  \caption{\textbf{Effectiveness of rolling-shutter compensation during initialization.}
  Colored dots indicate the estimated camera centers with respect to board (world) frame at keyframes. Circles show the 3D trajectories of camera centers, and arrows denote the normal vectors of planes from circle fitting. 
  Without compensation (a), the rolling-shutter camera (`Cam 1') exhibits a noticeably tilted normal vector.
  With compensation (b), the normal vectors of all camera trajectories are properly aligned.}
  \label{fig:RS_compensation}
  \vspace{-4mm}
\end{figure}

\noindent\textbf{Rolling-shutter Compensation}: \figref{fig:RS_compensation} qualitatively shows the effect of the rolling-shutter compensation module described in \secref{sec:preliminaries}.
We compare the camera-center trajectories computed from the camera-to-board poses $\pose{\mat{T}}{\coord{W}}{\coord{C}_i}\;(i=1,2,3)$ before and after optimization with rolling-shutter compensation.
Because all cameras are rigidly mounted on the turntable-driven platform, their trajectories should ideally share a common rotation axis, and the corresponding trajectory planes should be parallel.
Without compensation (a), the rolling-shutter camera (`Cam 1') exhibits a noticeably tilted trajectory plane.
With compensation (b), the fitted planes across three cameras become nearly parallel, indicating improved consistency with the expected rotational motion.
This supports the applicability of the proposed pipeline to practical multi-camera setups including rolling-shutter sensors.

\begin{figure}[t]
  \centering
  \includegraphics[width=1\linewidth]{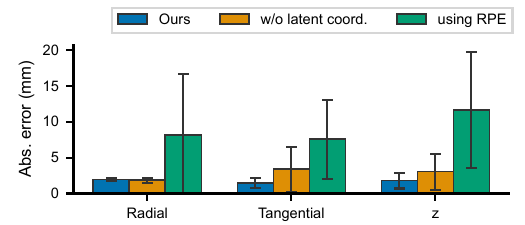}
  \vspace{-6mm}
  \caption{
  \textbf{Comparison of the proposed method against two design variants on the camera rig.}
  Bar plots report the \ac{MAE} of the component-wise translation errors.
  We compare our full method (`Ours') with two variants: one without the latent turntable frame (`w/o latent coord.') and one that uses reprojection error as the cost function (`using RPE').
  Error bars denote 1-sigma.}
  \label{fig:ablation_barplot}
  \vspace{-5mm}
\end{figure}

\noindent\textbf{Latent Frame and 3D Loss}: We also validate two key design choices of the proposed formulation on the camera rig.
Because rotation components do not show differences across variants, we report only translation results for clarity.

First, we consider a variant that removes the latent turntable frame (`w/o latent coord.'), where the turntable motion is estimated directly by extracting a screw motion from the camera-to-board measurements rather than being represented via $\coord{A}(t)$.
As shown in \figref{fig:ablation_barplot}, this variant produces noticeably larger translation errors, especially in the tangential and $z$ components.
This indicates that the latent frame effectively constrains ambiguities that arise when estimating camera motion directly in the world frame.

We further compare the proposed 3D error with a standard reprojection-error objective (`using RPE'). The reprojection-error variant shows larger translation errors and variance, with the largest degradation in the $z$ component (\figref{fig:ablation_barplot}).
This can be attributed to the fact that reprojection error is defined in the 2D image frame and does not explicitly enforce 3D rigid-body trajectory consistency under the turntable motion model, whereas the proposed 3D error does.

\subsection{Robustness under Non-Ideal Turntable Motion}
\label{sec:axis_deviation}

\begin{figure}[t]
  \centering
  \includegraphics[width=1\linewidth]{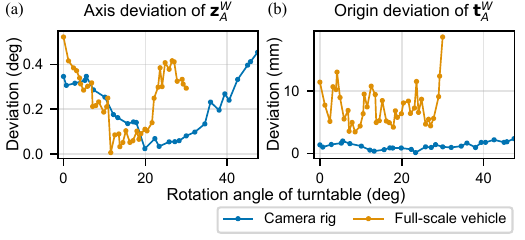}
  \vspace{-6.5mm}
  \caption{\textbf{Turntable motion deviations observed on the both platforms.} 
  Deviations over one revolution of the reference camera are plotted versus turntable angle, shown only over angles where the calibration board is visible in their \ac{FOV}.
  (a) \textit{Rotational deviations}: deviation of the turntable axis, defined as the direction of the $z$-axis of the $\coord{A}(t)$ expressed in $\coord{W}$.
  (b) \textit{Translational deviations}: deviation of the turntable origin from $\pose{\bvec{t}}{\coord{W}}{\coord{A}(0)}$.
  }
  \label{fig:axis_deviation}
  \vspace{-3mm}
\end{figure}

The ideal turntable assumptions in \secref{sec:preliminaries} are not perfectly met on real hardware due to wobble, vibration, and backlash\footnote{Here, \emph{wobble} refers to periodic oscillation of the rotation axis, \emph{vibration} to high-frequency fluctuations of the platform, and \emph{backlash} to lost motion caused by mechanical clearance in the drive.}.
In this section, we quantify these non-ideal motions on both the camera rig and the full-scale vehicle platform, and discuss their implications for calibration performance.

In \figref{fig:axis_deviation} (a), we plot the turntable-axis deviation, defined as the direction of the $z$-axis of the turntable frame $\coord{A}(t)$ expressed in $\coord{W}$, as a function of the turntable rotation angle.
While an ideal turntable would yield deviations close to zero, both platforms exhibit non-negligible axis wobble.
In \figref{fig:axis_deviation} (b), we show the turntable-origin deviation, quantified from $\pose{\bvec{t}}{\coord{W}}{\coord{A}(0)}$, which reveals millimeter-level fluctuations consistent with vibration and backlash.

Importantly, despite these non-ideal motions, the proposed method maintains the calibration accuracy reported in \secref{sec:testbed_eval} and \secref{sec:vehicle_eval}.
That is, the observed axis and origin deviations do not translate into divergent or unstable extrinsic estimates; instead, they are largely absorbed by the global optimization that estimates $\pose{\mat{T}}{\coord{A}}{\coord{C}}$ as a single time-invariant parameter over all keyframes.
Overall, these results support that the proposed approach is practically robust to turntable imperfections commonly present in cost-effective hardware.

\subsection{Comparison to Target-based Method in Overlapping FOV}
\label{sec:compare_overlap}

\begin{table}[!t]
\centering

\caption{Side-by-side comparison with a conventional target-based method and ours in two stereo configurations with overlapping \ac{FOV}.
Rotation is reported as (Roll, Pitch, Yaw) and translation as ($x, y, z$).
The last column reports the discrepancy between the two estimates as \ac{RRE} for rotation rows and \ac{RTE} for translation rows.
Units are in $\deg$ for rotation and $\mm$ for translation.}
\vspace{-2mm}
\label{tab:comparison_baseline}

\resizebox{\columnwidth}{!}{
\renewcommand{\arraystretch}{1.4}
\begin{tabular}{c|c|c|c|c}
\hline
\multicolumn{2}{c|}{\textbf{Setting}}
& \textbf{Target-based}
& \textbf{Ours} 
& \celltab{c}{c}{\textbf{Diff.} \\ \textbf{norm}} \\
\hline
\multirow{2}{*}{Stereo-A}
& Rot. 
& (-0.24, -0.33, 0.61)
& (-0.24, -0.33, 0.57)
& 0.06 \\
& Trans.
& (98.76, 1.18, -0.82)
& (98.96, 1.27, -0.88)
& 0.23 \\
\hline

\multirow{2}{*}{Stereo-B}
& Rot.
& (0.11, 0.63, -0.01)
& (0.11, 0.63, -0.04)
& 0.03   \\
& Trans.
& (98.30, -10.21, -1.46)
& (98.35, -10.21, -1.50)
& 0.07 \\
\hline
\end{tabular}
}
\vspace{-5mm}
\end{table}

As an additional reference, we compare the proposed method with a conventional target-based calibration method~\cite{zhang2000flexible} on two stereo configurations with overlapping \ac{FOV}s (Stereo-A and Stereo-B in \tabref{tab:comparison_baseline}).
We apply our method without modification (same formulation as the non-overlapping setting), whereas the conventional method exploits simultaneous board observations in the overlapping views.
As shown in \tabref{tab:comparison_baseline}, the two methods yield nearly matching estimates, differing by at most $0.1\degree$ in rotation and $0.3,\mm$ in translation.
Since our method is built on sequential board observations and a turntable kinematic model rather than direct inter-camera correspondences, it does not fundamentally depend on \ac{FOV} overlap and is expected to maintain similar accuracy in non-overlapping settings.
\section{Conclusion}
\label{sec:conclusion} 

In this paper, we presented a pure-rotation-based extrinsic calibration method for multi-camera systems with non-overlapping \ac{FOV}s, designed for practical deployment. 
By rotating a rigidly mounted multi-camera system in front of a calibration board, the proposed framework provides a space-efficient and low-burden calibration procedure that is well suited to industrial platforms where setup complexity and repeatability are critical.

The method introduces a latent frame and a 3D loss on $\SE3$, enabling global optimization from sequential target observations without direct inter-camera correspondences.
Experiments on a camera rig and a full-scale vehicle validated stable and repeatable calibration performance across diverse configurations, including rolling-shutter sensing and non-ideal turntable motion.
Overall, the results support the proposed method as a practical calibration solution that balances accuracy and usability for real multi-camera systems.

\small
\bibliographystyle{IEEEtranN} 
\bibliography{string-short,references}

\end{document}